\PassOptionsToPackage{unicode}{hyperref}
\PassOptionsToPackage{hyphens}{url}
\PassOptionsToPackage{dvipsnames,svgnames,x11names}{xcolor}
\documentclass[
  11pt,
]{article}
\usepackage{xcolor}
\usepackage[margin=1in]{geometry}
\usepackage{amsmath,amssymb}
\usepackage{iftex}
\ifPDFTeX
  \usepackage[T1]{fontenc}
  \usepackage[utf8]{inputenc}
  \usepackage{textcomp} 
\else 
  \usepackage{unicode-math} 
  \defaultfontfeatures{Scale=MatchLowercase}
  \defaultfontfeatures[\rmfamily]{Ligatures=TeX,Scale=1}
\fi
\usepackage{lmodern}
\ifPDFTeX\else
\fi
\IfFileExists{upquote.sty}{\usepackage{upquote}}{}
\IfFileExists{microtype.sty}{
  \usepackage[]{microtype}
  \UseMicrotypeSet[protrusion]{basicmath} 
}{}
\makeatletter
\@ifundefined{KOMAClassName}{
  \IfFileExists{parskip.sty}{%
    \usepackage{parskip}
  }{
    \setlength{\parindent}{0pt}
    \setlength{\parskip}{6pt plus 2pt minus 1pt}}
}{
  \KOMAoptions{parskip=half}}
\makeatother
\usepackage{color}
\usepackage{fancyvrb}

\DefineVerbatimEnvironment{Highlighting}{Verbatim}{commandchars=\\\{\}}
\newenvironment{Shaded}{}{}

\newcommand{\AttributeTok}[1]{\textcolor[rgb]{0.49,0.56,0.16}{#1}}

\newcommand{\CommentTok}[1]{\textcolor[rgb]{0.38,0.63,0.69}{\textit{#1}}}

\newcommand{\DataTypeTok}[1]{\textcolor[rgb]{0.56,0.13,0.00}{#1}}

\newcommand{\ExtensionTok}[1]{#1}

\newcommand{\NormalTok}[1]{#1}

\usepackage{longtable,booktabs,array}
\usepackage{calc} 
\usepackage{etoolbox}
\makeatletter
\patchcmd\longtable{\par}{\if@noskipsec\mbox{}\fi\par}{}{}
\makeatother
\IfFileExists{footnotehyper.sty}{\usepackage{footnotehyper}}{\usepackage{footnote}}
\makesavenoteenv{longtable}
\usepackage{graphicx}
\makeatletter
\newsavebox\pandoc@box
\newcommand*\pandocbounded[1]{
  \sbox\pandoc@box{#1}%
  \Gscale@div\@tempa{\textheight}{\dimexpr\ht\pandoc@box+\dp\pandoc@box\relax}%
  \Gscale@div\@tempb{\linewidth}{\wd\pandoc@box}%
  \ifdim\@tempb\p@<\@tempa\p@\let\@tempa\@tempb\fi
  \ifdim\@tempa\p@<\p@\scalebox{\@tempa}{\usebox\pandoc@box}%
  \else\usebox{\pandoc@box}%
  \fi%
}
\def\fps@figure{htbp}
\makeatother
\NewDocumentCommand\citeproctext{}{}

\makeatletter
 \let\@cite@ofmt\@firstofone
 \def\@biblabel#1{}
 \def\@cite#1#2{{#1\if@tempswa , #2\fi}}
\makeatother
\newlength{\cslhangindent}
\newlength{\csllabelwidth}
\newenvironment{CSLReferences}[2] 
 {\begin{list}{}{%
  \setlength{\itemindent}{0pt}
  \setlength{\leftmargin}{0pt}
  \setlength{\parsep}{0pt}
  \ifodd #1
   \setlength{\leftmargin}{\cslhangindent}
   \setlength{\itemindent}{-1\cslhangindent}
  \fi
  \setlength{\itemsep}{#2\baselineskip}}}
 {\end{list}}
\usepackage{calc}

\usepackage{bookmark}
\IfFileExists{xurl.sty}{\usepackage{xurl}}{} 
\hypersetup{
  pdftitle={Deterministic Integrity Gates for LLM-Assisted Clinical Manuscript Preparation: An Auditable Biomedical Informatics Architecture},
  pdfauthor={Yoojin Nama,b,c; Jinhoon Jeongd; Namkug Kima,b,*},
  colorlinks=true,
  linkcolor={blue},
  filecolor={Maroon},
  citecolor={Blue},
  urlcolor={Blue},
  pdfcreator={LaTeX via pandoc}}

\title{Deterministic Integrity Gates for LLM-Assisted Clinical
Manuscript Preparation: An Auditable Biomedical Informatics
Architecture}
\author{Yoojin Nam\textsuperscript{a,b,c} \and Jinhoon Jeong\textsuperscript{d} \and Namkug Kim\textsuperscript{a,b,*}}
\date{}

\begin{document}
\maketitle

\textsuperscript{a} Department of Radiology and Research Institute of
Radiology, University of Ulsan College of Medicine, Asan Medical Center,
Seoul, Republic of Korea \textsuperscript{b} Department of Convergence
Medicine, University of Ulsan College of Medicine, Asan Medical Center,
Seoul, Republic of Korea \textsuperscript{c} Aperivue, Incheon, Republic
of Korea \textsuperscript{d} Department of Medical Science, AMIST, Asan
Medical Center, University of Ulsan College of Medicine, Seoul, Republic
of Korea

* Corresponding author: Namkug Kim, Department of Radiology and
Department of Convergence Medicine, University of Ulsan College of
Medicine, Asan Medical Center, 88 Olympic-ro 43-gil, Songpa-gu, Seoul
05505, Republic of Korea. Email: namkugkim@gmail.com

\clearpage

\section*{Abstract}\label{abstract}
\addcontentsline{toc}{section}{Abstract}

\textbf{Objective.} As autonomous research agents and AI co-scientist
systems push large language models (LLMs) from drafting toward
end-to-end manuscript production, the bottleneck shifts from generation
to verification. In clinical research writing, fluent LLM output can
hide fabricated citations, numbers that drift from source tables, and
unmet reporting-guideline items. Existing tools generate text without
verifying it, and self-critique inherits the blind spots that produce
confident fabrication. We describe an architecture that pairs generation
with verification.

\textbf{Methods.} The design rests on three principles: decompose the
workflow into self-contained skills, gate every stage transition with
halt-on-failure, and resolve each integrity question with the cheapest
sufficient mechanism: a deterministic, re-executable check where one
suffices, and a prose-level probe only where interpretation is
unavoidable. This determinism-where-possible split, organized as an
integrity-gate taxonomy, is the core contribution. It is realized as
MedSci Skills, an open-source toolkit of 43 skills with a deterministic
tier of 21 standard-library detectors. We evaluate it on three
reproducible public-dataset pipelines (STARD, PRISMA, STROBE) and a
seeded-defect ablation.

\textbf{Results.} Across the three pipelines every content-hash manifest
verified clean and the gates surfaced real defects, including a
prognostic claim in a single-time-point study corrected to
association-only language. On 27 identical injected defects the
deterministic gates detected all 27 with no false positives on the
matched clean fixtures, whereas a generic single-prompt LLM reviewer
detected 11, its misses concentrated in generated-code,
bibliography-internal, and style defects the prose does not expose.

\textbf{Conclusion.} Determinism-where-possible verification yields an
auditable, re-executable trail that exposes the evidence a human needs
to check an LLM-assisted manuscript. This is feasibility and
reproducibility evidence, not a claim of human-competitive quality,
which a separate blinded study addresses. MedSci Skills is MIT-licensed
and archived (version 3.8.0).

\section*{Keywords}\label{keywords}
\addcontentsline{toc}{section}{Keywords}

Large language models; Research integrity; Reporting guidelines;
Reproducibility; Deterministic verification; Clinical research writing

\clearpage

\section{Introduction}\label{introduction}

Large language models (LLMs) are entering the clinical research-writing
workflow at speed, drafting sections, summarizing evidence, formatting
references, and assembling whole manuscripts; their multimodal
successors are advancing on the same trajectory across medical imaging
and practice {[}1{]}. Large-scale text analysis now suggests that
LLM-associated language is widespread in published academic articles,
making verification infrastructure relevant to routine scholarly
production rather than only to experimental AI-writing workflows
{[}2{]}. The appeal is obvious: a researcher can move from data to a
structured draft in hours rather than weeks. Yet the same fluency that
makes these models useful also makes their failures hard to see. Text
that reads as authoritative can carry fabricated citations, numbers that
drift between the tables and the prose, and silent gaps against the
reporting standards that journals require.

These failure modes are documented rather than hypothetical. Studies of
LLM-generated bibliographies report references that are plausibly
formatted but do not exist or do not support the citing sentence
{[}3{]}. Numerical inconsistency is subtler: a model regenerating a
paragraph may restate a sample size, an effect estimate, or a confidence
interval in a form that no longer matches the source table, with nothing
in the text to flag the discrepancy. And adherence to the reporting
guidelines curated by the EQUATOR Network {[}4{]} (among them STROBE,
STARD and its AI extension, TRIPOD+AI, CLAIM, and PRISMA and its
diagnostic-test-accuracy extension {[}5--7{]}) is something an LLM can
describe convincingly without satisfying item by item.

The tools that put LLMs to work in research writing largely
\emph{generate} text; they do not \emph{verify} it. A model can be
prompted to self-critique, and frameworks such as Self-Refine and
Reflexion show that iterative self-feedback improves output when the
flaw is recognizable from the text itself {[}8,9{]}. But self-correction
degrades when the error turns on an external fact the model does not
hold; LLMs have been shown unable to reliably self-correct reasoning
without external feedback {[}10{]}. Asking the model that wrote a
citation to judge whether the citation is real inherits its blind spots,
and a confident hallucination is exactly the case a self-check is least
likely to catch. What is missing is not more generation, and not more
LLM self-grading, but a layer that checks the claims a generated
manuscript makes against external authorities (a reference database, the
source data, a reporting checklist) using deterministic procedures
wherever one suffices, and reserving model judgment for the questions
that genuinely require it.

This paper describes such a layer and the architecture that makes it
auditable. We make three contributions. First, we present a design that
decomposes the medical-research writing workflow into self-contained,
composable skills, chains them through an explicit orchestrator, and
places a pass-gate at every stage transition so that a failure halts the
pipeline rather than propagating downstream. Second, we introduce an
integrity-gate taxonomy that separates checks a clean lookup can settle
(citation existence and author cross-checks against PubMed and CrossRef,
three-way reconciliation between source data, analysis output, and
prose, cohort and pool arithmetic, and reporting-checklist coverage)
from the prose-level probes that need a human or a model in the loop,
and we implement the deterministic checks as standard-library detectors
with synthetic fixtures and regression tests. Third, we release the
architecture as an open-source, version-pinned, archive-deposited
toolkit and demonstrate it on three reproducible public-dataset
pipelines, each carrying a content-hash lock so that the archived result
tables re-verify byte for byte and the reported values reproduce on
re-execution up to documented toolchain-version tolerance. We bound the
claim deliberately: the contribution is an architecture that produces an
\emph{auditable trail a human can verify}, not a system that replaces
human peer review or is shown here to match human writing quality.

\subsection{Problem}\label{problem}

LLMs accelerate clinical manuscript writing but can introduce fabricated
citations, numerical drift between tables and prose, and unmet
reporting-guideline items that fluent text conceals.

\subsection{What is already known}\label{what-is-already-known}

AI writing tools largely generate text without verifying it, and a model
asked to critique its own output inherits the blind spots that produce
confident fabrication.

\subsection{What this paper adds}\label{what-this-paper-adds}

An integrity-gated architecture that pairs generation with verification,
resolving each integrity question with the cheapest sufficient mechanism
(deterministic re-executable checks where possible, prose probes only
where interpretation is unavoidable), released as an open-source toolkit
whose deterministic gates caught all 27 seeded defects with no false
positives on the matched clean fixtures, versus 11 of 27 for a
single-prompt LLM reviewer.

\subsection{Who would benefit from the knowledge in this
paper}\label{who-would-benefit-from-the-knowledge-in-this-paper}

Clinical researchers, biomedical-informatics tool builders, and journal
editors and reviewers who need auditable, reproducible AI-assisted
manuscript preparation.

\section{Related Work}\label{related-work}

The credibility of a clinical research report depends less on its prose
than on whether it states what a reader needs in order to appraise it:
who was studied, how the index test was interpreted, what was adjusted
for, where the numbers came from. The EQUATOR Network curates this
expectation as reporting guidelines {[}4{]}: STROBE for observational
studies {[}11{]}, STARD for diagnostic accuracy {[}12{]}, PRISMA for
systematic reviews {[}13{]}, and TRIPOD for prediction models {[}14{]}.
As machine-learning methods entered clinical research, the community
extended them with AI-aware variants: TRIPOD+AI {[}5{]}, the STARD-AI
protocol {[}15{]}, PRISMA-DTA {[}7{]}, and CLAIM for medical-imaging AI
{[}6{]}. What matters here is their form. Each guideline is an
enumerated list of discrete, individually checkable items, so ``does
this manuscript address item N'' has a definite answer, a target for
automated checking categorically different from judging whether the
writing is good.

A second development concerns how capability is packaged for LLM agents.
Recent tooling has converged on the notion of a skill: a self-contained
unit bundling instructions, executable scripts, and reference assets
behind a name an agent can invoke, with an emerging cross-host
specification so the same skill runs under different runtimes {[}16{]}.
Decomposition into skills is not merely an engineering convenience; it
is what makes verification possible. A monolithic prompt that ingests
data and emits a finished manuscript offers no seam at which an
intermediate product can be inspected, whereas a chain of named skills
exposes a well-defined boundary at every transition (after the analysis,
after the figures, after the draft, after the compliance check), and
that boundary is exactly where a gate can sit.

A third line of work improves LLM output by having the model critique
and revise its own work, as in Self-Refine {[}8{]} and Reflexion
{[}9{]}. Iterative self-feedback helps when the error is recognizable
from the text alone, such as a logical gap or an unclear passage. It
helps far less for the errors that dominate clinical drafting, and
models have been shown unable to reliably self-correct without external
feedback {[}10{]}, because those errors turn on an external fact the
model does not hold: whether a DOI resolves to a real article, whether a
sample size in the prose matches the source table, whether item 14 of
STARD is genuinely satisfied.

The architecture sits against four adjacent lines of work a reader will
rightly compare it to. First, autonomous research agents now attempt the
full arc from idea to manuscript: end-to-end systems have produced
machine-learning papers submitted to, and in some cases accepted at,
workshop venues {[}17{]}, and multi-agent ``co-scientist'' systems
assist across hypothesis generation and experimental design {[}18{]},
including domain applications such as statistical genetics {[}19{]},
with the broader landscape reviewed elsewhere {[}20{]}. Such systems
maximize generation; the present work is orthogonal, supplying the
verification layer those pipelines lack rather than competing on
autonomy. A contemporaneous line shares this verification-first emphasis
from the machine-learning-theory side: a human-on-the-loop pipeline that
fixes the quality bar while automating each stage behind named on-disk
artifacts and stage-transition validation gates {[}21{]}, and a
companion framework that scales an agent's warranted autonomy to the
strength of its available verifier and cautions against artifacts that
appear settled before their underlying claims are checked {[}22{]}; the
present work instantiates that discipline for medical research, with
reporting-guideline and source-to-prose gates. The same
verification-first stance is being carried into open scientific
discovery, where an agent-native platform has multiple agents
collaborate against public, deterministic verifiers and exact arithmetic
makes correctness checkable at the task itself {[}23{]}; clinical
writing affords no comparable ready-made verifier, which is precisely
why the deterministic boundary here must be constructed per integrity
question rather than inherited from the problem. Second, fabricated or
inaccurate references in LLM-written text are now quantified in medicine
{[}24{]}, and a recent audit of 2.5 million PubMed Central Open Access
biomedical papers identified fabricated references at growing frequency
{[}25{]}, underscoring that citation verification has become a live
publishing-integrity problem rather than a hypothetical failure mode,
which is precisely what the citation-existence and author cross-check
gates target. Third, the automation of evidence-synthesis chores such as
risk-of-bias assessment has an established track record {[}26{]}; our
contribution is not one more single-purpose checker but the organizing
taxonomy that decides, per integrity question, whether a deterministic
check suffices. Fourth, the closest neighbors outside medicine are
retrieval-augmented generation with citation grounding and attribution
verification {[}27{]}, which check generated text against a supplied
corpus, and verifier or guardrail frameworks {[}28{]} that wrap a
generator with an external validator. The architecture differs from the
first in checking already-generated prose against external bibliographic
authorities such as PubMed and CrossRef and against the source data,
rather than grounding at generation against a provided corpus; and from
the second in supplying not a single validator but a taxonomy that
decides, per question, whether a deterministic check suffices, together
with domain-specific reporting-guideline and source-to-prose
reconciliation gates. The components it composes are individually
established; the contribution is their medical-research-specific
organization under the determinism-where-possible discipline, contrasted
with these neighbors capability by capability in Table 4.

These observations point to one design principle: retain the generative
strength of LLM skills, but verify each claim by the cheapest sufficient
mechanism. A deterministic procedure serves where the claim reduces to a
lookup or an arithmetic identity (a citation exists or it does not, a
row of counts sums to its total or it does not, a checklist item is
addressed or it is not); a prose-level probe, supplied by a human or by
a model acting as an external reviewer rather than as the author of the
text under review, serves only where the claim genuinely requires
interpretation. The architecture described next is organized around this
division of labor.

\section{Methods}\label{methods}

\subsection{Architecture}\label{architecture}

The architecture treats manuscript preparation as a pipeline of named
skills coordinated by a single orchestrator, with a verification gate
interposed at every transition between skills (Figure 2). It rests on
three principles: decomposition of the workflow into self-contained
skills, gating at every stage with halt-on-failure, and the use of
deterministic verification wherever a deterministic procedure suffices,
falling back to a prose-level probe only where interpretation is
genuinely required. The third principle is the paper's central claim,
and it is what distinguishes the design from an LLM that critiques its
own drafts.

\emph{Decomposition.} Each stage of the research-writing workflow is a
self-contained skill: literature search, statistical analysis, figure
generation, manuscript drafting, reporting-guideline compliance,
reference verification, and pre-submission self-review, among others. A
skill bundles its instructions, executable scripts, and reference assets
behind a name the orchestrator invokes, and it communicates with the
next skill through explicit, inspectable artifacts on disk rather than
hidden internal state. A single entry point routes a request to the
appropriate skill or chains several into a multi-step pipeline, and an
end-to-end mode runs the chain autonomously with post-skill validation.
Decomposition makes each intermediate product a first-class, inspectable
object and gives every stage transition a well-defined boundary at which
a gate can run.

\emph{Gating.} At each boundary the orchestrator runs the gate
appropriate to the stage and halts the pipeline on failure rather than
letting a defective artifact flow downstream. A failed citation audit
stops the manuscript before it is built into a submission package; a
numerical inconsistency between the analysis output and the prose stops
the draft before it reaches the compliance check. Halt-on-failure is
deliberate: in a generation pipeline the cost of a silent error
compounds, because each subsequent stage treats the flawed upstream
artifact as ground truth. By making the gate a precondition for
advancing, the architecture converts errors that would otherwise surface
at submission (or after publication) into early, localized failures with
a specific, re-runnable diagnostic.

\emph{Determinism where possible.} The gates are not uniform. The design
classifies every verification by the kind of judgment it requires and
resolves it with the cheapest sufficient mechanism. A large class of
integrity questions reduces to a lookup or an arithmetic identity, and
for these a deterministic checker is strictly preferable to model
judgment: it is reproducible, it is auditable, and it cannot itself
fabricate a verdict (though, as a false positive in the demonstrations
below shows, it can still encode a wrong assumption). Does a cited DOI
resolve in CrossRef and does its author list match the citing text? Does
every count in the prose reconcile with the source table and the
analysis script? Does an exclusion cascade sum to the analytic sample?
Is each applicable reporting-guideline item addressed? These are settled
by deterministic detectors that return the same verdict on every run and
leave a trace a reviewer can re-execute. Only the questions that
genuinely turn on interpretation (whether a limitation is candid,
whether a conclusion overreaches the design) are routed to a prose-level
probe supplied by a human or by a model acting as an external reviewer
rather than as the author of the text under review.

\emph{The integrity-gate taxonomy.} Organizing the gates by this
principle yields the taxonomy in Figure 1. The deterministic tier is
implemented as a set of standard-library detectors grouped into five
families: citation and reference integrity and adequacy (existence and
author cross-checks against PubMed and CrossRef, citation-key and
cross-reference validation, and a reference-adequacy check that every
named method is cited); numerical, cohort, and pool arithmetic
(three-way reconciliation between source data, analysis output, and
prose; exclusion-cascade and partition balance; figure-legend count
reconciliation; multi-copy divergence); confounding, scope, and estimand
contracts (an endpoint-to-conclusion scope gate that flags, for example,
a cross-sectional design licensing a prognostic claim; a structural-zero
guard; a confounding-completeness check); reporting compliance
(checklist coverage, base-plus-extension framework naming, flow-diagram
reconciliation); and style and review-process integrity (classical-style
body conventions, reviewer-team consistency, and a generated-code
quality gate that lints emitted analysis scripts for missing seeds,
hard-coded data, and non-portable paths). The deterministic tier is
deliberately conservative: a detector fires only when it can do so
without interpretation, so that a flag is informative rather than noise.
Where a check cannot be made deterministic without overreaching, it is
left as a prose probe rather than forced into a brittle heuristic. The
five families are not offered as an a priori or exhaustive partition:
they were distilled empirically from recurring integrity defects
observed across prior medical-writing projects and then implemented as
detectors, so the taxonomy is best read as a reference instantiation of
the determinism-where-possible discipline, the transferable
contribution, rather than a claim that these categories carve the space
completely.

\subsection{Implementation}\label{implementation}

We implement the architecture as MedSci Skills, an open-source toolkit
of 43 skills spanning the medical-research lifecycle from literature
search and study design through statistical analysis, figure generation,
manuscript drafting, reporting-guideline compliance, reference
verification, journal selection, and revision response. A single
orchestrator is the entry point: it classifies a request and routes it
to the appropriate skill, or chains several skills into a multi-step
pipeline, and an end-to-end mode executes the full chain autonomously
with post-skill validation and halt-on-failure at each transition. The
skills communicate through artifacts on disk, so the orchestrator's
gates inspect the same files a human would open.

The deterministic tier of the taxonomy is realized as 21 detectors that
use only the Python standard library, living inside the skills with no
third-party dependencies, most accompanied by privacy-free synthetic
fixtures and a regression test wired into continuous integration (the
exact coverage is reported in Table 3; a detector-by-family inventory is
given in the Supplementary Material). Restricting the detectors to the
standard library keeps them portable and auditable: a reviewer can read
a detector end to end and re-run it on any manuscript without installing
a toolchain. The stdlib constraint applies to the integrity detectors;
the demonstration analyses themselves use standard scientific packages
(scikit-learn, metafor, the R survey package), whose versions are pinned
in the archived release. The toolkit audits 32 EQUATOR reporting
guidelines and risk-of-bias instruments through the compliance skill,
and the catalog counts cited throughout the documentation (the number of
skills, guidelines, and detectors) are recomputed from disk and asserted
against a single source-of-truth file in continuous integration, so a
documentation count cannot silently drift from the artifacts it
describes. The toolkit thus applies its own integrity principle to
itself.

Reproducibility is enforced at the data layer. Each demonstration
pipeline carries a content-hash manifest recording a SHA-256 over every
input and derived table together with a per-column value hash, and a
verification command recomputes the hashes and reports any drift. A
demonstration that verifies clean re-verifies its archived tables byte
for byte, and re-executing it reproduces the hash-locked derived values
up to documented toolchain-version tolerance, so a manuscript built on
it inherits a checkable chain from raw data to reported number; the
manifest establishes stored-output integrity rather than guaranteeing
byte-identical re-execution across toolchains, following established
reproducible-research and FAIR software-citation principles {[}29,30{]}.
The skills are model- and host-portable: the same skills are documented
to run under Claude Code, OpenAI Codex, Cursor, and GitHub Copilot, with
portability checked by an installer self-test and a path-contract scan
rather than by live execution inside each host, so the architecture is
not bound to a single vendor's runtime.

The toolkit is released under the MIT license, version-pinned, and
archived with a citable digital object identifier. The work reported
here corresponds to a single frozen release: MedSci Skills version
3.8.0, under the concept identifier 10.5281/zenodo.20155321
(version-specific identifier 10.5281/zenodo.20582972, tag commit
60ce35c). The two-snapshot arrangement is deliberate and load-bearing
for reproducibility rather than incidental version bookkeeping. The
three demonstrations were generated on the v3.7.0 snapshot (version DOI
10.5281/zenodo.20577997, commit 5adda7c) and carried into v3.8.0
byte-unchanged; v3.8.0 adds only the evaluation harness and its
committed canonical run on top of those frozen demonstration artifacts.
A reader reproduces any demonstration number (Table 1) from the v3.7.0
archive and any evaluation number (Tables 2 and 3) from the v3.8.0
archive, in each case from a fixed artifact rather than from the moving
development branch; the version identifiers are the coordinates that let
a specific number be traced to the exact archived state that produced
it.

\subsection{Evaluation design}\label{evaluation-design}

We evaluate the toolkit in two complementary ways: three end-to-end
demonstrations that exercise the pipeline on real public datasets, and a
reproducible evaluation harness, archived with the release, that tests
the instrument itself. The harness analyses are instrument-level
(whether the detectors fire on the defects they target, whether the
artifacts regenerate, and whether the audit trail is machine-traceable)
and make no claim about the merit of the manuscripts the tool produces,
which a separate blinded study addresses. All harnesses reported here
are deterministic code executed by the first author; each returns the
same verdict on every run and is re-executable by any reader from the
archived release, so the triggering and exact-match results below carry
no human judgment. The few points where a judgment is unavoidable are
confined and recorded, with per-claim records archived.

For the demonstrations, we regenerated three complete manuscripts from
scratch on the frozen v3.7.0 release, each from a different public
dataset and study type, and each targeting a different reporting
guideline, to span the dominant study-type families in clinical research
(diagnostic accuracy, evidence synthesis, and observational
epidemiology), so that the demonstration exercises three distinct
reporting guidelines (STARD, PRISMA, STROBE) and the distinct gate paths
they trigger. We drew each from a canonical, openly redistributable
public dataset because a worked demonstration of reproducibility
requires that a reader be able to re-run it and re-verify every number;
reconstructing inputs from the tables of recently published medical-AI
papers, an alternative we considered, would both constrain
redistribution and risk training-data contamination that confounds a
demonstration of language-model generation, so we reserved that
extension for future work. Each pipeline read only its canonical
open-source data and was driven by the v3.7.0 skills, so the run is an
independent test of the released toolkit rather than a replay.

The seeded-defect harness takes the clean demonstration artifacts (and,
for citation and Methods-to-Results checks, a small synthetic fixture,
because the demonstration bibliographies are deliberately empty
placeholders), injects exactly one known defect at a time into a
temporary copy, runs the targeted detector, and records whether it
emitted the specific code for that defect; each detector is also run on
the clean input to measure false positives. The 27 offline injection
instances come from 17 distinct injectors, several applied across all
three demonstration manuscripts; two further citation injectors need a
live PubMed or CrossRef lookup (a fabricated DOI and a wrong author
list) and are recorded as not-run, for 19 distinct injectors in total.
Because fault injection has no defined defect prevalence, we report
recall and the clean false-positive rate, not precision. Each injected
defect instantiates a documented real-world error class (a fabricated,
duplicate, or placeholder citation; an unbalanced exclusion cascade; a
cross-sectional design licensing a prognostic claim; an AI-extension
framework named without its base instrument; a machine-written style
marker; or a non-reproducible analysis script), with the rationale,
provenance, and completeness scope documented in a defect-rationale file
archived with the release. The set is deliberately family-complete and
grounded in recurring failure modes; we do not claim it is exhaustive,
and it is best read as a regression-style challenge suite for known
integrity-failure modes rather than a representative sample of
real-world manuscript errors. To isolate what the deterministic tier
contributes, we ran an ablation that removes it and substitutes a
generic single-prompt LLM review on the identical 27 defects, using one
current model from the family that generated the demonstrations, one
fixed prompt, a single sample, and provider-default sampling on one
date. The remaining harnesses probe reproducibility, auditability, and
portability: fresh-clone manifest verification, an audit-trail probe
classifying each claim by whether it exact-matches a manifest-locked
table value, an installer self-test and path-contract scan, a detector
coverage inventory, and an injected-metadata-drift check on the
catalog-consistency validator.

\section{Results}\label{results}

\subsection{Demonstrations}\label{demonstrations}

The three pipelines ran end to end, and every content-hash manifest
verified clean on an independent re-run, so each demonstration's
archived tables re-verify byte for byte (Table 1). The headline analyses
behaved as their datasets dictate. On the Wisconsin Breast Cancer
dataset (Demonstration 1, STARD 2015), all three machine-learning index
tests were highly discriminative (logistic-regression area under the
curve 0.998, 95\% CI 0.994--1.000) with no pairwise difference detected
(all three DeLong comparisons p \textgreater{} 0.5); because the index
tests are machine-learning models, the STARD-AI protocol {[}15{]}, at
protocol stage when these demonstrations were prepared, is named
alongside the finalized base STARD. The BCG meta-analysis (Demonstration
2, PRISMA 2020) reproduced the canonical pooled estimate (risk ratio
0.49, 95\% CI 0.34--0.70; I² = 92.2\%; 95\% prediction interval
0.15--1.55), which is its own reproducibility check. The NHANES
2017--2018 cross-sectional study (Demonstration 3, STROBE) found obesity
associated with diabetes after survey-weighted adjustment (adjusted odds
ratio 3.03, 95\% CI 2.29--4.02). The internal checklist assessment, an
item-by-item self-assessment rather than an external reporting-quality
judgment, ranged from 60.9\% to 83.3\% of applicable items (mean 68.7\%
± 12.7\%, n = 3, descriptive only), reporting genuine gaps rather than
inflating coverage; demonstration body length averaged 1,576 ± 209
words.

The feasibility claim rests less on those summary numbers than on the
gates visibly doing their job. The generated-code gate flagged a
hard-coded absolute path and two dead imports in the Demonstration 1
analysis script, corrected before the script was accepted. The
scope-coherence gate flagged prognostic vocabulary in the conclusion of
the cross-sectional Demonstration 3 (language a single-time-point design
cannot support), and the manuscript was reworded to association-only
claims and re-verified clean; this firing is shown end to end as a
worked example in Supplementary Figure S1. The classical-style gate
caught a decimal inconsistency in Demonstration 2. Whether each flag was
a genuine defect or a false positive was judged by the first author
against the source artifact, and because every flag is a re-runnable
diagnostic pointing at a specific line or value, a reader can confirm
the same judgment independently rather than take it on trust. The trail
also records an honest false positive: in Demonstration 3 the
confounding-completeness check fired because its heuristic assumes an
exposure-stratified baseline table whereas this study tabulated by
outcome, and the actual confounder was in the adjustment set. We report
it rather than suppress it, because a verification layer earns trust by
being inspectable, including where it is wrong.

We are explicit about what these demonstrations do and do not establish.
They establish that the pipeline runs end to end across three study
types, that its outputs are reproducible from archived artifacts, and
that its integrity gates surface real defects with re-executable
evidence. They do not establish that the resulting manuscripts are of
publishable quality or competitive with human writing: the self-review
verdicts span ``pass,'' ``revise,'' and ``accept with notes'' (Table 1),
and the references in every demonstration were left as unverified
placeholders by design, precisely to exercise the reference-adequacy
gate. Establishing quality is the work of a separate, blinded
evaluation. The contribution here is the auditable trail, not a verdict
on the prose it accompanies.

\textbf{Table 1. Three artifact-clean demonstrations regenerated on
MedSci Skills v3.7.0.}

{\def\LTcaptype{none} 
\begin{longtable}[]{@{}
  >{\raggedright\arraybackslash}p{(\linewidth - 6\tabcolsep) * \real{0.2200}}
  >{\raggedright\arraybackslash}p{(\linewidth - 6\tabcolsep) * \real{0.2600}}
  >{\raggedright\arraybackslash}p{(\linewidth - 6\tabcolsep) * \real{0.2600}}
  >{\raggedright\arraybackslash}p{(\linewidth - 6\tabcolsep) * \real{0.2600}}@{}}
\toprule\noalign{}
\begin{minipage}[b]{\linewidth}\raggedright
\end{minipage} & \begin{minipage}[b]{\linewidth}\raggedright
Demonstration 1
\end{minipage} & \begin{minipage}[b]{\linewidth}\raggedright
Demonstration 2
\end{minipage} & \begin{minipage}[b]{\linewidth}\raggedright
Demonstration 3
\end{minipage} \\
\midrule\noalign{}
\endhead
\bottomrule\noalign{}
\endlastfoot
\textbf{Dataset (source)} & Wisconsin Breast Cancer (sklearn) & BCG
vaccine trials (metafor dat.bcg) & NHANES 2017--2018 (CDC) \\
\textbf{Sample} & N = 569 (212 malignant) & k = 13 trials, N = 357,347 &
N = 5,010 adults \\
\textbf{Design} & Diagnostic accuracy & Intervention meta-analysis &
Cross-sectional, survey-weighted \\
\textbf{Reporting guideline} & STARD 2015 & PRISMA 2020 & STROBE \\
\textbf{Headline result} & LR AUC 0.998 (0.994--1.000); no pairwise AUC
difference detected (DeLong p \textgreater{} 0.5) & Pooled RR 0.49
(0.34--0.70); I² 92.2\% & Adjusted OR 3.03 (2.29--4.02) \\
\textbf{Reporting compliance (applicable)} & 14/23 = 60.9\% & 24/42 =
57.1\% (→ 26/42 = 61.9\% after fix) & 25/30 = 83.3\% \\
\textbf{Self-review (initial → final)} & 82 → 88 (PASS) & 78 → 82
(REVISE) & REVISE → ACCEPT-WITH-NOTES \\
\textbf{Reproducibility lock} & verify --strict 5/5 & verify --strict
4/4 & verify --strict 6/6 \\
\end{longtable}
}

\emph{All values are pulled from the frozen v3.7.0 run and independently
re-verified (detector re-runs, manifest verification, source-table
cross-checks). The self-review score and verdict are the toolkit's own
internal grader output, not an external measure of manuscript quality;
they are reported only to show the fix loop ran, and the companion
study, not this score, evaluates quality. Reporting-compliance
denominators exclude items not applicable to a public-dataset
computational demonstration. The Demonstration 2 arrow shows the in-loop
change after the fix added a registration statement; Demonstrations 1
and 3 compliance was unchanged by the fix loop. Across the three
demonstrations, applicable-item reporting compliance averaged 68.7\%
(range 60.9--83.3\%; SD 12.7, using each demonstration's post-fix value)
and body length averaged 1,576 words (range 1,340--1,738; SD 209), with
n = 3; because the three are deliberately different study types reported
against different guidelines, these summaries are descriptive only and
not a pooled performance estimate.}

\subsection{Seeded-defect detection and
ablation}\label{seeded-defect-detection-and-ablation}

Figure 3 summarizes the evaluation harness. Across 27 offline injection
instances spanning all five gate families, the deterministic detectors
recovered every injected defect and raised none of the corresponding
signals on the clean inputs (Table 2 and Figure 3, panel A). To isolate
the deterministic tier's contribution, the ablation presented the same
27 defects to a generic single-prompt LLM review. Asked to list every
problem it could find (the same model family used to generate the
demonstrations, so the comparison also instances the self-review case in
which the model that produced a text is asked to find its own faults),
the model recovered 11 of 27 (41\%), against 27 of 27 (100\%) for the
deterministic detectors. The shortfall is not uniform. The model did
reasonably on prose-visible semantic issues: a conclusion overreaching a
cross-sectional design, an AI-extension framework named without its
base, an in-prose arithmetic contradiction, a promised-but-absent
analysis, and a visibly undefined citation key. But it systematically
missed the classes a generated manuscript hides from a prose reviewer.
In the style and review-process family, which aggregates nine
style-convention markers and five generated-code defects, it caught only
two: one style marker (a section symbol) and one code defect (a missing
seed). The in-body AI-disclosure sentence and em-dash overuse are
conventions a generic reviewer does not treat as errors; the remaining
code defects (a hard-coded path, an in-place overwrite) live in the
analysis script rather than the prose; and it missed both
bibliography-internal citation defects (a placeholder pagination and a
duplicated entry in the reference file), neither of which appears in the
prose. Part of this gap is definitional: the generated-code and
bibliography-internal classes cannot surface in the manuscript text a
generic reviewer reads, so a prose-only reviewer cannot reach them by
construction, and even on the prose-visible classes the gap persists.
This is the result the design anticipates: the deterministic tier earns
its place precisely on integrity questions that turn on an external
fact, a machine-checkable convention, or an artifact the prose does not
expose. We frame this strictly as a recall comparison under one model,
one fixed prompt, and one date, not as a universal model-superiority
claim and not as a judgment of manuscript quality. A second
LLM-dependent harness, a bounded self-review convergence loop with
logging fields adapted from MI-CLEAR-LLM {[}31{]}, is specified and
archived with the release but was not executed for this release.

\textbf{Table 2. Seeded-defect detection by gate family: deterministic
gates versus a generic single-prompt LLM reviewer (offline benchmark,
identical defects).}

{\def\LTcaptype{none} 
\begin{longtable}[]{@{}
  >{\raggedright\arraybackslash}p{(\linewidth - 8\tabcolsep) * \real{0.3200}}
  >{\raggedright\arraybackslash}p{(\linewidth - 8\tabcolsep) * \real{0.1500}}
  >{\raggedright\arraybackslash}p{(\linewidth - 8\tabcolsep) * \real{0.1900}}
  >{\raggedright\arraybackslash}p{(\linewidth - 8\tabcolsep) * \real{0.1600}}
  >{\raggedright\arraybackslash}p{(\linewidth - 8\tabcolsep) * \real{0.1800}}@{}}
\toprule\noalign{}
\begin{minipage}[b]{\linewidth}\raggedright
Gate family
\end{minipage} & \begin{minipage}[b]{\linewidth}\raggedright
Defects injected
\end{minipage} & \begin{minipage}[b]{\linewidth}\raggedright
Deterministic gates (recall)
\end{minipage} & \begin{minipage}[b]{\linewidth}\raggedright
Clean false positives
\end{minipage} & \begin{minipage}[b]{\linewidth}\raggedright
Generic LLM reviewer (recall)
\end{minipage} \\
\midrule\noalign{}
\endhead
\bottomrule\noalign{}
\endlastfoot
\textbf{Citation and reference integrity} & 3 & 3 (100\%) & 0 & 1
(33\%) \\
\textbf{Numerical, cohort, and pool arithmetic} & 3 & 3 (100\%) & 0 & 2
(67\%) \\
\textbf{Confounding, scope, and estimand contracts} & 1 & 1 (100\%) & 0
& 1 (100\%) \\
\textbf{Reporting compliance} & 6 & 6 (100\%) & 0 & 5 (83\%) \\
\textbf{Style and review-process integrity} & 14 & 14 (100\%) & 0 & 2
(14\%) \\
\textbf{Total} & \textbf{27} & \textbf{27 (100\%)} & \textbf{0} &
\textbf{11 (41\%)} \\
\end{longtable}
}

\emph{One defect injected per temporary copy; one target detector judged
per defect, so attribution is unambiguous. The 27 offline injection
instances come from 17 distinct injectors, several applied across all
three demonstration manuscripts; two further network-required citation
injectors (fabricated DOI, wrong author) were not run, for 19 distinct
injectors in total. Recall is detected over injected; it measures
detector triggering, not population sensitivity. Clean false positives
count only the target defect signal raised on a clean input; an expected
baseline signal, such as the offline reference checker marking every
entry unverified, is not counted. The final column is the ablation: the
same 27 defects presented to a generic single-prompt LLM review (one
model, one fixed prompt, one date), recovering 11 of 27 (41\%), with the
gap concentrated in the style/review-process, generated-code, and
bibliography-internal classes the prose does not expose. The ``Style and
review-process integrity'' row aggregates style-convention and
generated-code detectors (nine and five injected defects), and
``Numerical, cohort, and pool arithmetic'' includes the
artifact-coverage detector; the per-injector breakdown is in the
archived defect-rationale file and the Supplementary Material.}

\subsection{Reproducibility, audit-trail, and
portability}\label{reproducibility-audit-trail-and-portability}

A fresh checkout of the release regenerated all three demonstration
manifests cleanly under independent verification (Table 3 and Figure 3,
panel B). An audit-trail probe extracted 249 candidate claims from the
three demonstration manuscripts, of which 136 were numerical; of those,
74 (54\%) resolved by exact match to a value in a manifest-locked
analysis table, 48 (35\%) were partial matches (a rounded or derived
restatement, which is still a trace), and 14 (10\%) did not trace to a
locked table value. The claim extraction and the exact-match comparison
are run by a deterministic script, so the 54\% figure reproduces on
re-execution; only the partial-versus-adjudicated split involves
documented human judgment, and the per-claim records are archived. The
54\% is a floor on machine-checkability for the claim type where exact
tracing is meaningful, not a quality score. The installer self-test
confirmed that all skills are discoverable from the converged install
directories without touching a real host directory, a path-contract scan
found no hard-coded personal paths in skill logic, and the four
documented host targets were confirmed against the compatibility matrix.
A coverage inventory enumerated the deterministic detectors and recorded
which ship a synthetic fixture and a regression test. Finally, injecting
metadata drift into a temporary copy of the repository confirmed that
the catalog-consistency validator catches each drift.

\textbf{Table 3. Reproducibility, audit-trail, and portability checks.}

{\def\LTcaptype{none} 
\begin{longtable}[]{@{}
  >{\raggedright\arraybackslash}p{(\linewidth - 2\tabcolsep) * \real{0.2600}}
  >{\raggedright\arraybackslash}p{(\linewidth - 2\tabcolsep) * \real{0.7400}}@{}}
\toprule\noalign{}
\begin{minipage}[b]{\linewidth}\raggedright
Check
\end{minipage} & \begin{minipage}[b]{\linewidth}\raggedright
Result
\end{minipage} \\
\midrule\noalign{}
\endhead
\bottomrule\noalign{}
\endlastfoot
\textbf{Fresh-clone reproducibility} & 3/3 demonstration manifests
verify clean on an independent checkout \\
\textbf{Audit-trail completeness} & 249 manuscript claims extracted; of
136 numerical claims, 74 (54\%) exact-match a manifest-locked analysis
table, 48 (35\%) partial, 14 (10\%) untraced; numerical audit fully
deterministic \\
\textbf{Host portability} & installer self-test passes (all 43 skills
discoverable, no real host directory touched); 0 personal-path leaks in
skill logic; 4/4 documented host targets checked by installer self-test
and path-contract scan (not live in-host execution) \\
\textbf{Gate coverage inventory} & 21 deterministic detectors; 18 with a
regression test, 17 with a synthetic fixture \\
\textbf{Catalog drift resistance} & 4/4 injected metadata drifts caught
by the catalog-consistency validator \\
\end{longtable}
}

\emph{Each harness writes a self-describing run package (per-component
determinism class, input and output content hashes, exact commands, and
limitations) archived with the release; the deterministic harnesses
operate only on temporary copies and never modify the demonstration
artifacts or the repository. The coverage inventory is reported
honestly: the detectors that currently lack a synthetic fixture or a
regression test fall partly in the citation family, and closing this gap
is tracked future work.}

\section{Discussion}\label{discussion}

Biomedical research writing is itself an informatics problem: a
manuscript is the point where claims, source data, analysis code,
references, and reporting-guideline requirements must stay synchronized,
and a language model that raises drafting throughput also breaks that
synchronization in ways the prose does not reveal. The architecture
treats manuscript generation as a governed information pipeline rather
than a text-generation task, and its value is governance. In a workflow
where a language model can produce fluent error, the operative question
is not whether the model is ever wrong but whether a human can find out
before the error reaches a reader. Deterministic gates answer that
question for a defined class of failures: they make the failure visible
early, attach a re-executable diagnostic to it, and refuse to let the
defective artifact advance, while the content-hash lock carries a
checkable chain from raw data through analysis to the reported number.
The demonstrations show this is not aspirational; across three study
types the gates surfaced real defects as localized, inspectable flags
rather than as problems a reader would have to discover unaided.

Against the adjacent lines of work, the architecture's distinctive
element is the verification layer, not the generation it wraps. Generic
drafting and self-feedback methods optimize the text;
retrieval-augmented generation grounds claims against a supplied corpus;
verifier and guardrail frameworks wrap a generator with a
general-purpose validator; autonomous research agents maximize
end-to-end production. None supplies a medical-research-specific
verification layer that reconciles prose against the source data,
cross-checks citations against external bibliographic authorities, and
gates reporting-guideline coverage, with halt-on-failure between stages
(Table 4). The contribution is the per-question determinism taxonomy
that organizes these checks for the medical-research workflow, not any
single check in isolation.

\textbf{Table 4. How the verification layer compares with adjacent
approaches.}

{\def\LTcaptype{none} 
\begin{longtable}[]{@{}
  >{\raggedright\arraybackslash}p{(\linewidth - 12\tabcolsep) * \real{0.2800}}
  >{\raggedright\arraybackslash}p{(\linewidth - 12\tabcolsep) * \real{0.1200}}
  >{\raggedright\arraybackslash}p{(\linewidth - 12\tabcolsep) * \real{0.1200}}
  >{\raggedright\arraybackslash}p{(\linewidth - 12\tabcolsep) * \real{0.1200}}
  >{\raggedright\arraybackslash}p{(\linewidth - 12\tabcolsep) * \real{0.1200}}
  >{\raggedright\arraybackslash}p{(\linewidth - 12\tabcolsep) * \real{0.1200}}
  >{\raggedright\arraybackslash}p{(\linewidth - 12\tabcolsep) * \real{0.1200}}@{}}
\toprule\noalign{}
\begin{minipage}[b]{\linewidth}\raggedright
Verification capability
\end{minipage} & \begin{minipage}[b]{\linewidth}\raggedright
Generic LLM drafting
\end{minipage} & \begin{minipage}[b]{\linewidth}\raggedright
Self-Refine / Reflexion
\end{minipage} & \begin{minipage}[b]{\linewidth}\raggedright
RAG + citation grounding
\end{minipage} & \begin{minipage}[b]{\linewidth}\raggedright
Guardrail / verifier frameworks
\end{minipage} & \begin{minipage}[b]{\linewidth}\raggedright
Autonomous research agents
\end{minipage} & \begin{minipage}[b]{\linewidth}\raggedright
MedSci Skills
\end{minipage} \\
\midrule\noalign{}
\endhead
\bottomrule\noalign{}
\endlastfoot
\textbf{Source-to-prose numerical reconciliation} & Not primary target &
Not primary target & Not primary target & Not primary target & Not
primary target & Implemented \\
\textbf{Citation existence + author cross-check vs external authority
(PubMed/CrossRef)} & Not primary target & Not primary target & Related
capability & Related capability & Not primary target & Implemented \\
\textbf{EQUATOR reporting-guideline gating} & Not primary target & Not
primary target & Not primary target & Not primary target & Not primary
target & Implemented \\
\textbf{Cohort / pool arithmetic checks} & Not primary target & Not
primary target & Not primary target & Not primary target & Not primary
target & Implemented \\
\textbf{Content-hash data manifest (reproducibility lock)} & Not primary
target & Not primary target & Not primary target & Not primary target &
Related capability & Implemented \\
\textbf{Halt-on-failure stage orchestration} & Not primary target & Not
primary target & Not primary target & Related capability & Not primary
target & Implemented \\
\textbf{Medical-research reporting demonstration} & Not primary target &
Not primary target & Not primary target & Not primary target & Related
capability & Implemented \\
\end{longtable}
}

\emph{A conceptual capability comparison of the verification layer, not
an experimental ranking and not an overall-quality ranking; the
neighbours optimize generation or general-purpose validation and are
largely orthogonal. Entries summarize the primary design target of each
class as described in its cited source(s): ``Not primary target'' means
the capability is not described as a primary verification target of that
class, not that it could not be implemented as an extension; ``Related
capability'' marks a related but non-equivalent capability;
``Implemented'' means realized as a deterministic gate family exercised
in this work. The neighbour entries summarize self-feedback methods
{[}8,9{]}, retrieval-augmented generation with citation grounding
{[}27{]}, verifier/guardrail frameworks {[}28{]}, and autonomous
research agents {[}17{]}, discussed with their references in Related
Work; the MedSci Skills column corresponds to the deterministic gate
families exercised in this work. MedSci Skills is not claimed to be
globally superior; the claim is limited to the verification/governance
layer.}

These comparisons also clarify the relevance of the architecture to AI
co-scientist systems. In clinical research, the relevant question is not
only what an AI agent can generate, but which parts of its output can be
trusted, audited, and corrected before they enter the scientific record.
Recent AI scientist and co-scientist systems emphasize hypothesis
generation, multi-agent critique, experimental planning, and end-to-end
production {[}17--19{]}; MedSci Skills addresses a narrower,
complementary problem: the downstream governance of generated research
artifacts. Its pattern is to use generation where generation is useful,
but to verify with deterministic, re-executable checks whenever an
integrity question reduces to an external lookup, source-artifact
reconciliation, reporting requirement, or arithmetic identity. This
clinical orientation is consistent with broader concerns that medical AI
should be integrated in ways that preserve human calibration and
foundational expertise rather than bypass them {[}32{]}.

Several limitations bound these claims. First, the toolkit was primarily
authored by one of us, and we applied its own quality checks to this
manuscript; that dogfooding is a deliberate test of the architecture,
not an independent evaluation of it. The distinction matters at the
level of individual results: the detectors' triggering verdicts are
deterministic and operator-independent, but the human-judged labels
layered on top (whether a given firing is a true defect or a false
positive, and which demonstration outputs were judged clean before
injection) were adjudicated by the developer-author, so those judgments,
unlike the re-executable triggering results, most need the independent
replication described below. Second, the skill set reflects one
physician-researcher's domains (radiology, diagnostic-accuracy and
observational designs, and systematic review and meta-analysis), so its
coverage and defaults are correspondingly biased, and a researcher from
another field would find gaps. Third, this paper is itself a
software-and-methods report, a genre the toolkit's clinical-manuscript
machinery does not target: there is no STARD or STROBE checklist to
apply to an architecture paper, and dogfooding the toolkit on this
manuscript surfaced two instructive false positives: the scope-coherence
detector flags this paper because it repeatedly discusses prognostic
claims in a cross-sectional setting while describing what the gate
catches, and the reference-adequacy detector flags it because its
citation-count target assumes an original-research article type. Both
share one root cause, the absence of a software-and-methods article-type
profile, and supplying one is future work; we report these flags rather
than silence them, for the same reason we report the confounding one.
Fourth, the evidence here is feasibility by demonstration; external,
independent use of the toolkit by other researchers is untested, and a
powered assessment of detector precision and of the architecture in
other hands is the central direction for future work. The seeded-defect
benchmark measures recall on known injected defects (every
offline-injected defect was recovered) but does not measure the
false-negative rate on real manuscripts carrying naturally-occurring
errors the detectors were never built to anticipate; quantifying that,
like estimating detector precision, would require a corpus with
independently adjudicated ground-truth errors that this controlled
injection does not provide. Concretely, the planned next step is a
blinded, artifact-level evaluation in which independent reviewers
classify each gate firing as a true defect, a false alarm, or a
non-actionable warning, and compare time-to-detection and defect-repair
completeness against an ungated language-model-assisted workflow.

Deterministic verification narrows the space in which a human must look;
it does not remove the human. The correct reading of a clean gate is
that a specific mechanical failure mode did not occur, not that the
manuscript is correct. Our claim that the trail is one ``a human can
verify'' is, accordingly, a claim about re-executability rather than
reviewer behaviour: every flag is a re-runnable diagnostic and every
reported number carries a content-hash chain back to the source, so any
reader can independently reproduce the verification. Whether possessing
this trail measurably improves what a human reviewer catches (the
blinded reader test the claim invites) is a separate empirical question,
and it is precisely what the companion study is designed to answer.

The intended use is concrete. The primary operator is an
author-researcher running a pre-submission self-check, or a
research-integrity or editorial-office screener triaging an LLM-assisted
submission; the decision the trail informs is local and binary: whether
to advance an artifact or halt it and route a specific, re-runnable flag
to a human. This matters now because LLM-assisted manuscripts are
reaching journals faster than the deterministic integrity screening that
would catch their characteristic failures, and the gates target exactly
those failures at the point in the workflow where a human still has time
to act on a localized diagnostic. The toolkit is an authoring and
quality-control aid, not a clinical decision tool: it makes no
diagnostic or treatment recommendation, and its outputs require expert
review before any use. More broadly, the architecture is not specific to
medicine. Decomposition into gated skills, with determinism wherever a
deterministic procedure suffices, applies to any writing task checkable
against external authorities, from a legal brief verified against a
citator to a data-science report reconciled against its notebooks.
Independent work outside medicine reinforces the principle: a
gate-checked decomposition workflow for systems engineering, evaluated
across several language models, outperformed a single-pass baseline
{[}33{]}. The transferable contribution is the taxonomy itself: the
discipline of asking, for each integrity question, whether a
deterministic procedure will settle it before reaching for model
judgment.

\section{Conclusion}\label{conclusion}

We have presented an architecture for language-model-assisted medical
research writing whose premise is that generation must be paired with
verification, and whose design resolves each integrity question with the
cheapest sufficient mechanism: a deterministic, re-executable check
wherever one suffices, and a prose-level probe only where interpretation
is unavoidable. Implemented as an open-source toolkit of 43 skills with
21 standard-library integrity detectors, and demonstrated on three
reproducible public-dataset pipelines, the architecture produces an
auditable, re-executable trail that exposes the evidence needed for
human verification. Its applicability is immediate and practical: an
author-researcher or an editorial-office screener can use it to decide,
for an LLM-assisted manuscript, whether to advance an artifact or halt
it and route a specific re-runnable flag to a human, and the
determinism-where-possible discipline transfers to any writing task that
can be checked against external authorities. That auditable trail,
rather than any claim to replace human judgment or to match human
writing quality, is the contribution.

\section*{CRediT authorship contribution
statement}\label{credit-authorship-contribution-statement}
\addcontentsline{toc}{section}{CRediT authorship contribution statement}

\textbf{Yoojin Nam:} Conceptualization, Software, Methodology,
Investigation, Writing -- original draft, Writing -- review \& editing.
\textbf{Jinhoon Jeong:} Validation, Methodology, Writing -- review \&
editing. \textbf{Namkug Kim:} Supervision, Conceptualization, Writing --
review \& editing.

\section*{Declaration of generative AI and AI-assisted technologies in
the manuscript preparation
process}\label{declaration-of-generative-ai-and-ai-assisted-technologies-in-the-manuscript-preparation-process}
\addcontentsline{toc}{section}{Declaration of generative AI and
AI-assisted technologies in the manuscript preparation process}

During the preparation of this work the authors used Anthropic's Claude
(Opus 4.8 model family), accessed via the Claude Code agent runtime,
during May--June 2026, for two purposes: (i) to draft and revise the
text of this manuscript, and (ii) to generate the three demonstration
pipelines (analysis code, figures, and demonstration manuscripts) under
human direction. After using this tool, the authors reviewed and edited
the content as needed, verified all AI-assisted outputs against source
data and against the toolkit's deterministic integrity gates, and take
full responsibility for the content of the published article. Generative
AI was not used to fabricate or alter data or results; every reported
quantitative value derives from executed, version-pinned code, and the
verification trail is published with the archived release. A detailed
account of which steps were deterministic and which were generative, and
of how the demonstrations were generated (model identification,
prompting via versioned skills, execution, and
training-data-contamination considerations), is given in the
Supplementary Material and in the repository's PROVENANCE record.

\section*{Acknowledgements}\label{acknowledgements}
\addcontentsline{toc}{section}{Acknowledgements}

The authors thank the maintainers of the open-source datasets and
software used in the demonstrations (the scikit-learn project, the
metafor package, the R survey package, and the U.S. National Health and
Nutrition Examination Survey).

\section*{Funding}\label{funding}
\addcontentsline{toc}{section}{Funding}

This work received no specific funding from any agency in the public,
commercial, or not-for-profit sectors.

\section*{Declaration of competing
interest}\label{declaration-of-competing-interest}
\addcontentsline{toc}{section}{Declaration of competing interest}

Y.N. is the founder of Aperivue (Incheon, Republic of Korea), under
which the MedSci Skills toolkit described in this work is developed and
released, and is its principal developer and maintainer. Because this
paper evaluates the authors' own instrument, the authors disclose a
non-financial conflict of interest (an intellectual and reputational
interest, and a potential future-commercial interest, in the toolkit's
adoption) and do not claim to be disinterested evaluators of it. The
conflict is mitigated by design: every deterministic result reported
here is re-executable and independently reproducible from the archived
release, and an independent, blinded evaluation of the toolkit's outputs
is deferred to a separate study. Aperivue is presently an open-source
initiative; no paid product, licensing, or consulting revenue is
currently derived from MedSci Skills. The authors declare no other
competing interests.

\section*{Data and code availability}\label{data-and-code-availability}
\addcontentsline{toc}{section}{Data and code availability}

MedSci Skills is open-source (MIT) at
https://github.com/Aperivue/medsci-skills and archived on Zenodo
(version 3.8.0; concept DOI 10.5281/zenodo.20155321, which always
resolves to the latest version; v3.8.0 version DOI
10.5281/zenodo.20582972, tag commit 60ce35c). The three demonstrations
were generated on the v3.7.0 snapshot (version DOI
10.5281/zenodo.20577997; commit 5adda7c) and are archived unchanged in
v3.8.0, together with the evaluation harness, its canonical run, and the
content-hash manifests. The demonstrations comprise data-preparation and
analysis scripts, manuscripts, quality-control artifacts, and
content-hash manifests; citation metadata is provided in the
repository's CITATION.cff. All datasets used are public (scikit-learn,
the metafor dat.bcg set, and CDC NHANES 2017--2018).

\section*{References}\label{references}
\addcontentsline{toc}{section}{References}

\protect\phantomsection\label{refs}

\clearpage

\section*{Figures}\label{figures}
\addcontentsline{toc}{section}{Figures}

\includegraphics[width=0.8\linewidth,height=\textheight,keepaspectratio]{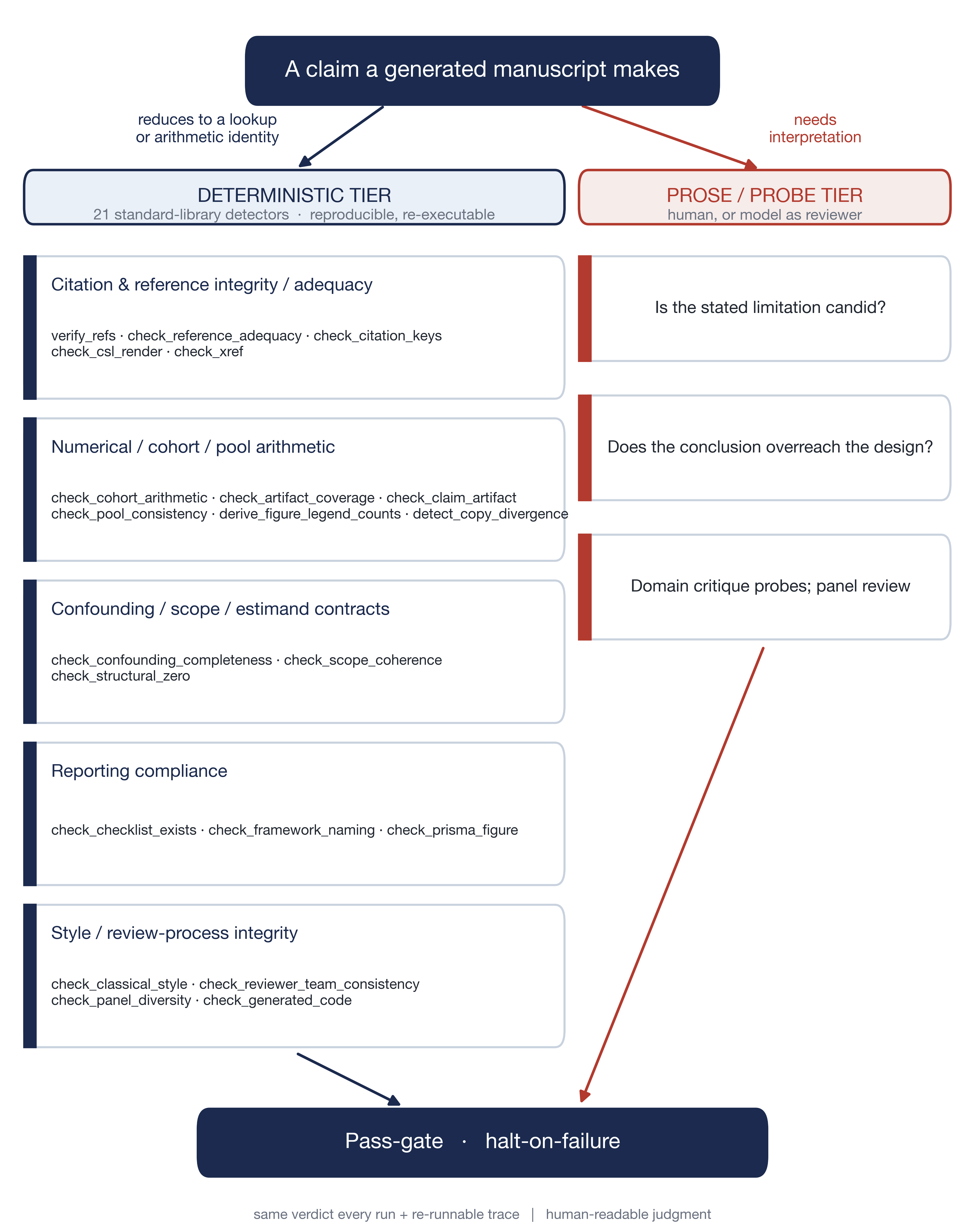}

\textbf{Figure 1. Integrity-gate taxonomy.} Each integrity question a
generated manuscript raises is routed by whether it reduces to a lookup
or arithmetic identity (the deterministic tier of 21 standard-library
detectors, grouped into five families) or whether it needs
interpretation (the prose/probe tier), with a pass-gate enforcing
halt-on-failure at the stage boundary.

\clearpage

\includegraphics[width=0.8\linewidth,height=\textheight,keepaspectratio]{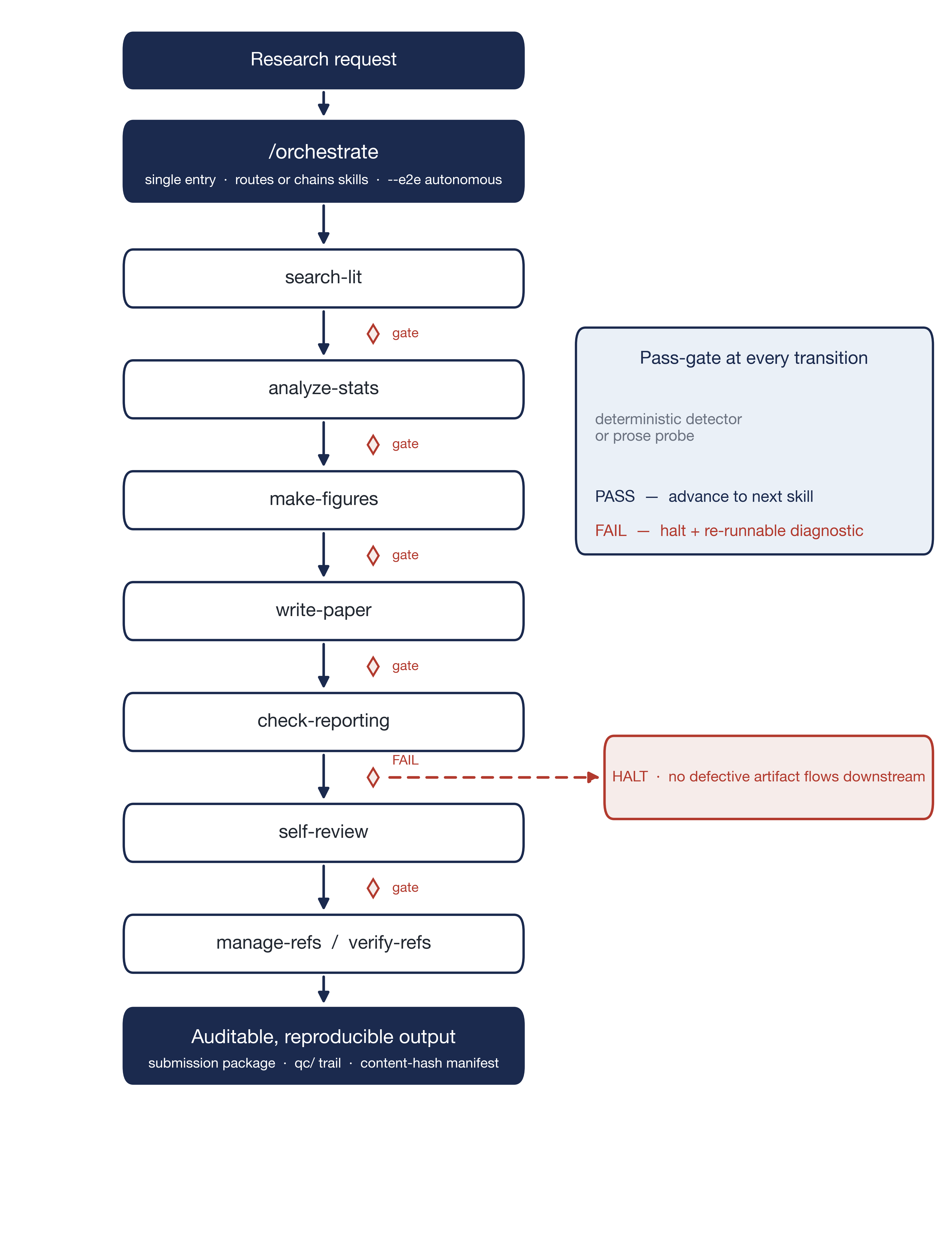}

\textbf{Figure 2. Orchestrated, gated pipeline.} A single orchestrator
routes a request or chains skills end to end, with a verification gate
at every transition; a passing gate advances the artifact while a
failing gate halts the pipeline and emits a re-runnable diagnostic,
yielding an auditable, reproducible output.

\clearpage

\includegraphics[width=0.85\linewidth,height=\textheight,keepaspectratio]{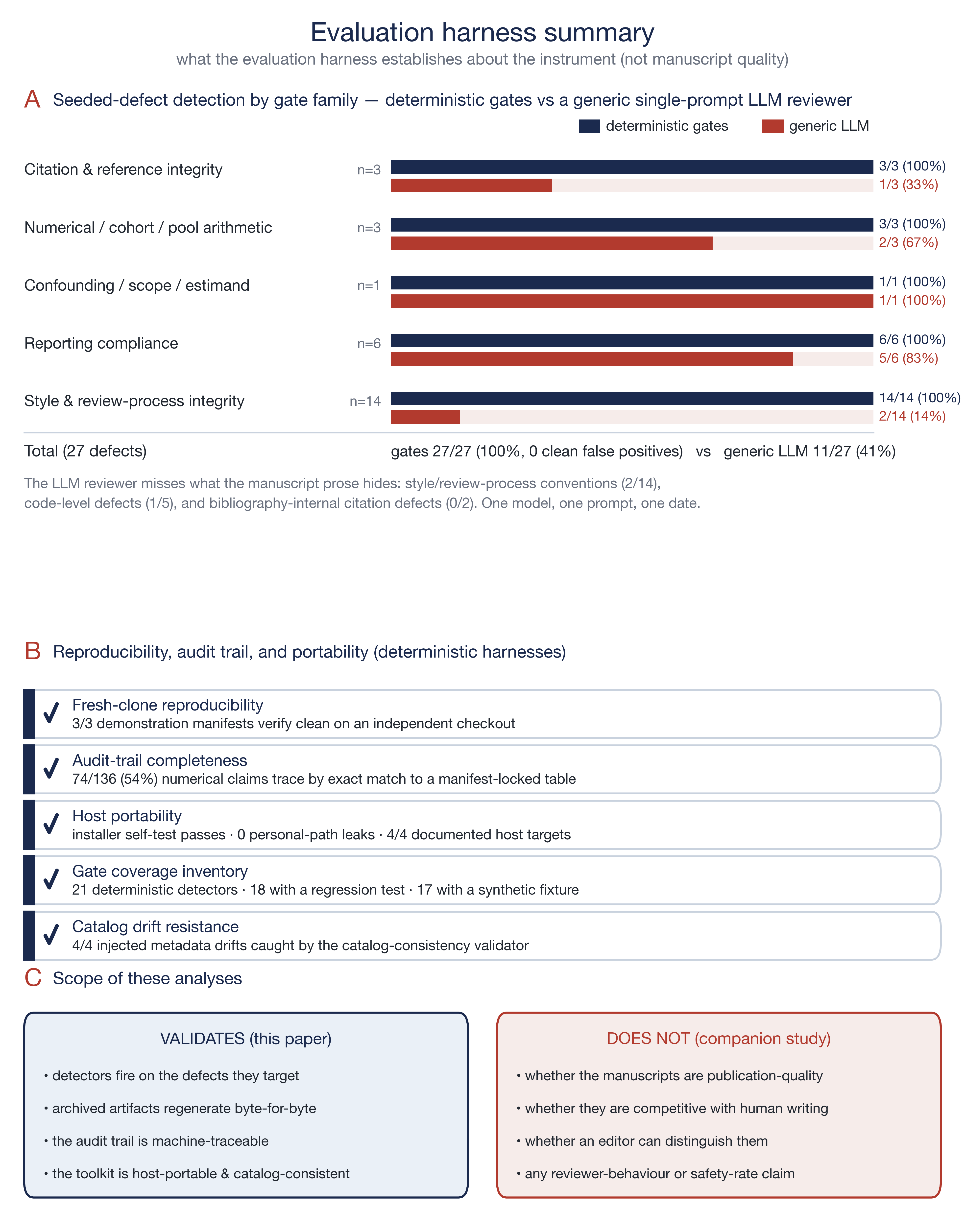}

\textbf{Figure 3. Evaluation-harness summary.} What the evaluation
harness establishes about the instrument, not about manuscript quality.
\textbf{(A)} Seeded-defect detection by gate family: the deterministic
gates recover all 27 injected defects with no clean false positives,
while a generic single-prompt LLM reviewer on the identical defects
recovers 11 of 27 (41\%), missing the style/review-process,
generated-code, and bibliography-internal classes the prose does not
expose (the ablation; full counts in Table 2). \textbf{(B)}
Reproducibility, audit-trail, and portability checks (full counts in
Table 3). \textbf{(C)} What these analyses do and do not establish, the
latter being the remit of the blinded companion study.

\section*{Supplementary Material}\label{supplementary-material}
\addcontentsline{toc}{section}{Supplementary Material}

This material supplements the main manuscript with the full
deterministic-detector inventory, the generation-transparency report for
the demonstrations, the per-class defect breakdown for the seeded-defect
benchmark, and the worked example of a caught defect. All numbers are
reproduced from the archived canonical run of MedSci Skills v3.8.0 (the
demonstrations from the v3.7.0 snapshot carried unchanged into v3.8.0).

\subsection*{S1. Deterministic-detector
inventory}\label{s1.-deterministic-detector-inventory}
\addcontentsline{toc}{subsection}{S1. Deterministic-detector inventory}

The deterministic tier of the integrity-gate taxonomy comprises 21
standard-library detectors (no third-party dependencies) living inside
the skills, grouped into the five families of the taxonomy (Figure 1 of
the main text). Table S1 lists them by family and source file.

\textbf{Table S1. The 21 deterministic detectors by gate family.}

{\def\LTcaptype{none} 
\begin{longtable}[]{@{}
  >{\raggedright\arraybackslash}p{(\linewidth - 2\tabcolsep) * \real{0.3000}}
  >{\raggedright\arraybackslash}p{(\linewidth - 2\tabcolsep) * \real{0.7000}}@{}}
\toprule\noalign{}
\begin{minipage}[b]{\linewidth}\raggedright
Gate family
\end{minipage} & \begin{minipage}[b]{\linewidth}\raggedright
Detectors (source file)
\end{minipage} \\
\midrule\noalign{}
\endhead
\bottomrule\noalign{}
\endlastfoot
\textbf{Citation and reference integrity and adequacy (5)} & DOI/PMID
and first/all-author cross-check (\texttt{verify-refs/verify\_refs.py});
reference adequacy, that every named method is cited
(\texttt{self-review/check\_reference\_adequacy.py}); citation-key
validation (\texttt{manage-refs/check\_citation\_keys.py}); CSL render
check (\texttt{manage-refs/check\_csl\_render.py}); cross-reference
validation (\texttt{manage-refs/check\_xref.py}) \\
\textbf{Numerical, cohort, and pool arithmetic (6)} &
cohort/exclusion-cascade arithmetic
(\texttt{self-review/check\_cohort\_arithmetic.py}); artifact coverage
(\texttt{self-review/check\_artifact\_coverage.py}); claim-to-artifact
reconciliation (\texttt{self-review/check\_claim\_artifact.py}); pool
consistency (\texttt{meta-analysis/check\_pool\_consistency.py});
figure-legend count reconciliation
(\texttt{make-figures/derive\_figure\_legend\_counts.py}); multi-copy
divergence (\texttt{sync-submission/detect\_copy\_divergence.py}) \\
\textbf{Confounding, scope, and estimand contracts (3)} & confounding
completeness (\texttt{self-review/check\_confounding\_completeness.py});
endpoint-to-conclusion scope coherence
(\texttt{self-review/check\_scope\_coherence.py}); structural-zero guard
(\texttt{clean-data/check\_structural\_zero.py}) \\
\textbf{Reporting compliance (3)} & checklist coverage
(\texttt{check-reporting/check\_checklist\_exists.py});
base-plus-extension framework naming
(\texttt{check-reporting/check\_framework\_naming.py}); PRISMA
flow-diagram reconciliation
(\texttt{check-reporting/check\_prisma\_figure.py}) \\
\textbf{Style and review-process integrity (4)} & classical-style body
conventions (\texttt{self-review/check\_classical\_style.py});
reviewer-team consistency
(\texttt{self-review/check\_reviewer\_team\_consistency.py}); panel
diversity (\texttt{self-review/check\_panel\_diversity.py});
generated-code quality
(\texttt{analyze-stats/check\_generated\_code.py}) \\
\end{longtable}
}

Of the 21 detectors, 18 ship with a regression test and 17 with a
privacy-free synthetic fixture wired into continuous integration (Table
3 of the main text); the detectors that currently lack a fixture or a
regression test fall partly in the citation family, and closing this gap
is tracked future work. The catalog counts cited throughout the
toolkit's documentation, namely the number of skills (43), reporting
guidelines and risk-of-bias instruments (32), and deterministic
detectors (21), are recomputed from disk and asserted against a single
source-of-truth file in continuous integration, so a documentation count
cannot silently drift from the artifacts it describes.

\subsection*{S2. Generation transparency for the
demonstrations}\label{s2.-generation-transparency-for-the-demonstrations}
\addcontentsline{toc}{subsection}{S2. Generation transparency for the
demonstrations}

For the three demonstrations we report the generation process following
the spirit of MI-CLEAR-LLM, a minimum-reporting standard for
large-language-model studies in healthcare. That standard is scoped to
studies evaluating LLM accuracy rather than to manuscript-writing
assistance, so we adopt its items as the closest-fit transparency
framework for an LLM-authored pipeline rather than as a compliance
target. The model is identified by family and version (Anthropic Claude,
Opus 4.8 model family) and accessed through a documented runtime (the
Claude Code agent runtime) over a stated date range (May--June 2026);
the prompts are the versioned skills themselves, published in the
archived release rather than paraphrased; each demonstration was
generated in an independent, artifact-clean session that read only its
canonical public dataset; and the resulting outputs were parsed and
checked by the deterministic gates. Because the demonstration datasets
are public, they may lie within the model's training corpus, so we make
no contamination-controlled accuracy claim; a companion study evaluates
contamination directly. The full quality-control trail of each
demonstration (the per-stage pipeline log, the literal detector outputs,
the reporting-checklist assessment, and the self-review record) is
included in the archived release alongside the analysis scripts and the
content-hash manifest, so a reader can re-execute every deterministic
check independently. The raw generation transcripts are retained and
available from the authors on request.

\subsection*{S3. Per-class breakdown of the seeded-defect
ablation}\label{s3.-per-class-breakdown-of-the-seeded-defect-ablation}
\addcontentsline{toc}{subsection}{S3. Per-class breakdown of the
seeded-defect ablation}

Table 2 of the main text reports detection by the five gate families.
Table S2 gives the finer per-class breakdown of the generic
single-prompt LLM reviewer's recall on the identical 27 injected
defects, which clarifies where the shortfall concentrates: the
style-convention and generated-code classes (which a prose reviewer
largely cannot see) and the bibliography-internal citation defects
(which do not appear in the prose).

\textbf{Table S2. Generic single-prompt LLM reviewer recall by defect
class (detected / injected).}

{\def\LTcaptype{none} 
\begin{longtable}[]{@{}ll@{}}
\toprule\noalign{}
Defect class & Detected / injected \\
\midrule\noalign{}
\endhead
\bottomrule\noalign{}
\endlastfoot
\textbf{Style conventions} & 1 / 9 \\
\textbf{Generated code} & 1 / 5 \\
\textbf{Citation (incl.~bibliography-internal)} & 1 / 3 \\
\textbf{Cohort / numerical} & 1 / 2 \\
\textbf{Reporting-framework naming} & 5 / 6 \\
\textbf{Scope / estimand} & 1 / 1 \\
\textbf{Artifact coverage} & 1 / 1 \\
\textbf{Total} & \textbf{11 / 27} \\
\end{longtable}
}

The deterministic gates detected 27 / 27 across these same classes with
no false positives on the clean inputs. The full per-injector rationale,
provenance, and completeness scope are documented in the
defect-rationale file archived with the release.

\subsection*{S4. Reproducing the two load-bearing
checks}\label{s4.-reproducing-the-two-load-bearing-checks}
\addcontentsline{toc}{subsection}{S4. Reproducing the two load-bearing
checks}

A reader can reproduce the two load-bearing checks from a clone of the
release with only the standard library plus \texttt{pandas}:

\begin{Shaded}
\begin{Highlighting}[]
\CommentTok{\# (1) re{-}verify one demonstration\textquotesingle{}s content{-}hash manifest}
\ExtensionTok{python3}\NormalTok{ skills/version{-}dataset/scripts/version\_dataset.py verify }\DataTypeTok{\textbackslash{}}
  \AttributeTok{{-}{-}manifest}\NormalTok{ demo/01\_wisconsin\_bc/manifest.lock.json }\DataTypeTok{\textbackslash{}}
  \AttributeTok{{-}{-}base}\NormalTok{ demo/01\_wisconsin\_bc }\AttributeTok{{-}{-}strict}

\CommentTok{\# (2) re{-}run the offline seeded{-}defect trigger benchmark (deterministic)}
\ExtensionTok{python3}\NormalTok{ evaluation/h1\_seeded\_defects/run\_h1.py}
\end{Highlighting}
\end{Shaded}

The first command exits zero when every file and per-column hash
matches; the second regenerates Table 2 and re-asserts its fixed
reproducibility hash. Demonstrations were generated under the package
versions recorded in the archived release; minor patch-version drift in
the analysis toolchain (for example, the R interpreter) does not affect
the derived values, which are fixed by the content-hash manifests.

\subsection*{S5. Worked example of a caught
defect}\label{s5.-worked-example-of-a-caught-defect}
\addcontentsline{toc}{subsection}{S5. Worked example of a caught defect}

\includegraphics[width=0.85\linewidth,height=\textheight,keepaspectratio]{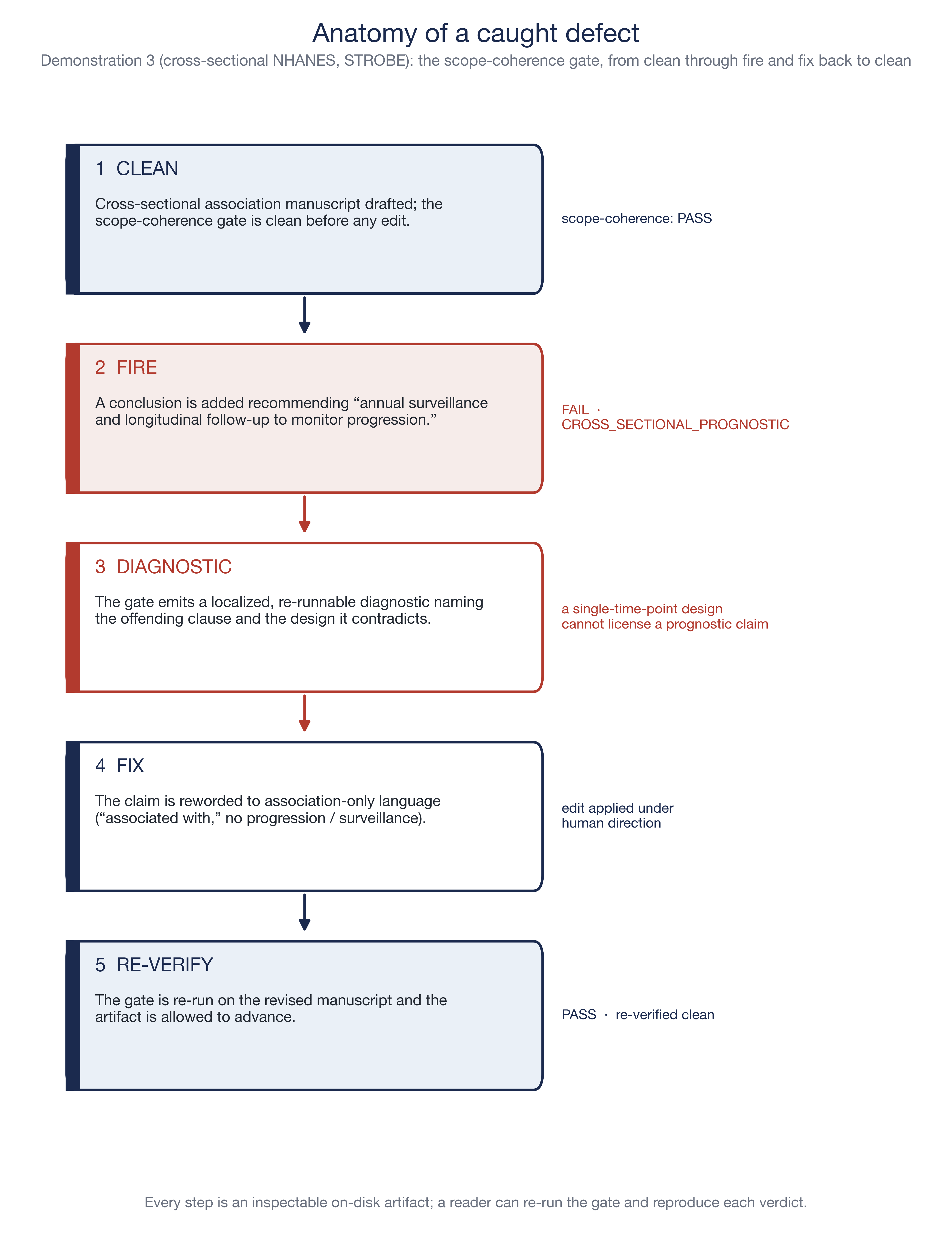}

\textbf{Supplementary Figure S1. Anatomy of a caught defect (worked
example).} The organic scope-coherence firing in the cross-sectional
Demonstration 3, shown as an inspectable trail: the gate is clean on the
drafted manuscript (1); fires when the conclusion is given prognostic
and surveillance language a single-time-point design cannot license (2,
code \texttt{CROSS\_SECTIONAL\_PROGNOSTIC}); emits a localized
re-runnable diagnostic naming the offending clause (3); the claim is
reworded to association-only language under human direction (4); and the
gate is re-run and the artifact re-verified clean (5). Every step is an
on-disk artifact a reader can re-execute.

\end{document}